\documentclass[letterpaper]{article} 
\usepackage{aaai2026}  
\usepackage{times}  
\usepackage{helvet}  
\usepackage{courier}  
\usepackage[hyphens]{url}  
\usepackage{graphicx} 
\usepackage{natbib}  
\usepackage{caption} 
\usepackage{algorithm}
\usepackage{algpseudocode}

\usepackage{enumitem}

\usepackage[dvipsnames]{xcolor}
\RequirePackage[colorlinks,citecolor=RoyalBlue]{hyperref}

\usepackage{amsmath}
\usepackage{array}
\usepackage{booktabs}
\usepackage{multirow}
\usepackage{cleveref}
\usepackage{graphicx}
\usepackage{subcaption}
\usepackage{booktabs}

\usepackage{amsthm}

\newtheoremstyle{nodecimal} 
  {3pt} 
  {3pt} 
  {\itshape} 
  {} 
  {\bfseries} 
  {} 
  {.5em} 
  {} 

\theoremstyle{nodecimal}

\newtheorem{proposition}{Proposition}

\newtheorem{assumption}{Assumption}
\newtheorem{remark}{Remark}

\usepackage{amsfonts}
\usepackage{amssymb}
\usepackage{amsmath}
\usepackage{fix-cm}
\DeclareMathOperator*{\argmin}{arg\,min}
\usepackage{newfloat}
\usepackage{listings}
\DeclareCaptionStyle{ruled}{labelfont=normalfont,labelsep=colon,strut=off} 
\floatstyle{ruled}
\newfloat{listing}{tb}{lst}{}
\floatname{listing}{Listing}
\title{Investigating Data Pruning for Pretraining Biological Foundation Models at Scale}
\author{
    Yifan Wu\textsuperscript{\rm 1,2,}\thanks{These authors contributed equally to this work.},
    Jiyue Jiang\textsuperscript{\rm 1,$\ast$,}\thanks{Corresponding authors.}, 
    Xichen Ye\textsuperscript{\rm 2}, 
    Yiqi Wang\textsuperscript{\rm 2}, 
    Chang Zhou\textsuperscript{\rm 1}, 
    Yitao Xu\textsuperscript{\rm 1}, 
    Jiayang Chen\textsuperscript{\rm 1}, \\
    He Hu\textsuperscript{\rm 3}, 
    Weizhong Zhang\textsuperscript{\rm 2,$\dagger$}, 
    Cheng Jin\textsuperscript{\rm 2}, 
    Jiao Yuan\textsuperscript{\rm 4,5,$\dagger$}, 
    Yu Li\textsuperscript{\rm 1,$\dagger$}
}
\affiliations{
    \textsuperscript{\rm 1}The Chinese University of Hong Kong\\
    \textsuperscript{\rm 2}Fudan University\\
    \textsuperscript{\rm 3}Guangdong Laboratory of Artificial Intelligence and Digital Economy (SZ)\\
    \textsuperscript{\rm 4}Guangzhou National Laboratory\\
    \textsuperscript{\rm 5}Guangzhou Medical University\\
    \{victorwu, jiangjy\}@link.cuhk.edu.hk, weizhongzhang@fudan.edu.cn, yuan\_jiao@gzlab.ac.cn,
    liyu@cse.cuhk.edu.hk
}

\usepackage{bibentry}

\begin{document}

\maketitle

\begin{abstract}
Biological foundation models (BioFMs), pretrained on large-scale biological sequences, have recently shown strong potential in providing meaningful representations for diverse downstream bioinformatics tasks. However, such models often rely on millions to billions of training sequences and billions of parameters, resulting in prohibitive computational costs and significant barriers to reproducibility and accessibility—particularly for academic labs. 
To address these challenges, we investigate the feasibility of data pruning for BioFM pretraining and propose a post-hoc influence-guided data pruning framework tailored to biological domains. 
Our approach first introduces a subset-based self-influence formulation that enables efficient estimation of sample importance at low computational cost. Built upon this, we propose two simple yet effective selection strategies: Top-$k$ Influence (Top I) and Coverage-Centric Influence (CCI).
Then, we empirically validate our method on two representative BioFMs: RNA-FM and ESM-C. 
For RNA, our framework consistently outperforms random selection baselines under an extreme pruning rate of over 99\%, which displays our framework's effectiveness.
Furthermore, we demonstrate the generalizability of our framework on protein-related tasks using ESM-C.
In specific, our coreset even outperforms random $10\times$ subsets in both RNA and protein settings, revealing substantial redundancy in biological sequence dataset.
These findings underscore the potential of influence-guided data pruning to substantially reduce the computational cost of BioFM pretraining, paving the way for more efficient, accessible, and sustainable biological AI research.
\end{abstract}

\begin{links}
  \link{Code}{https://github.com/victor-yifanwu/bio-coreset}
\end{links}

\section{Introduction}

\begin{figure*}[t]
    \begin{center}
    \centerline{\includegraphics[width=0.85\textwidth]{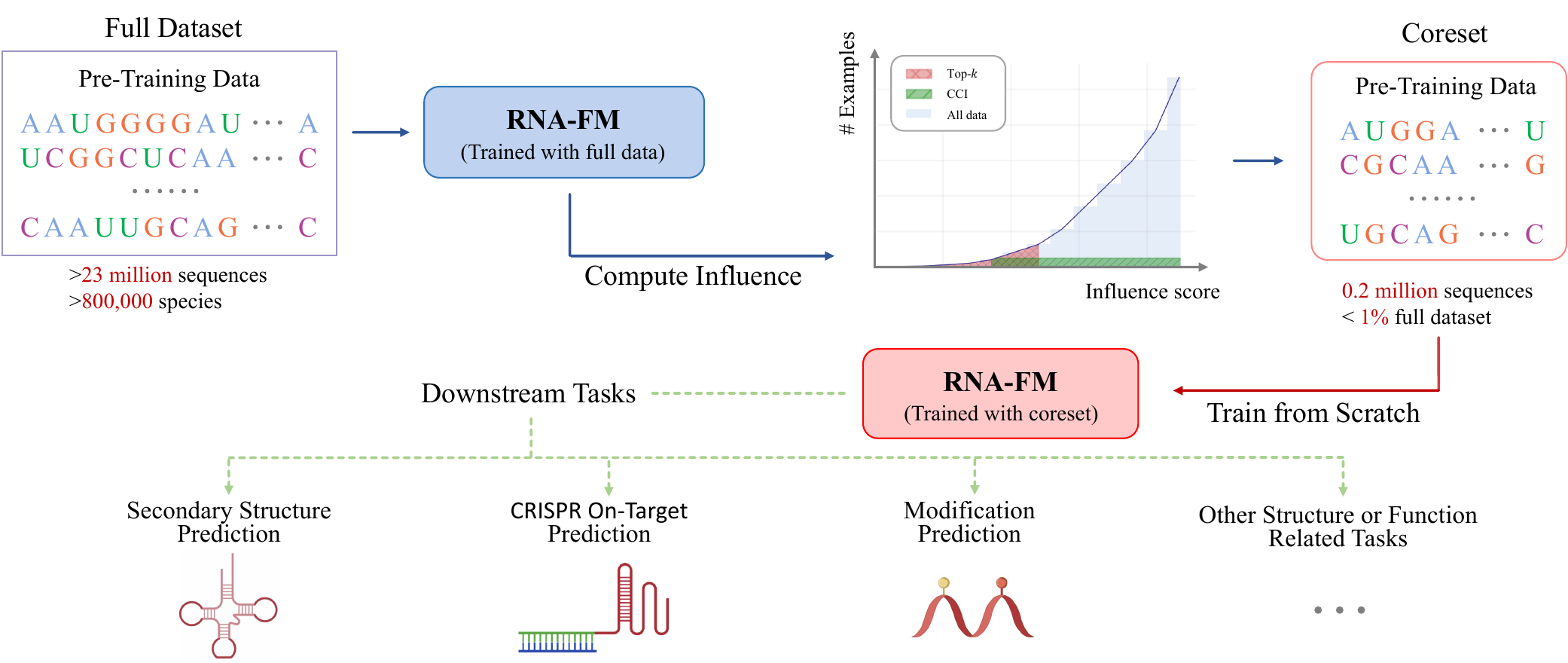}}
    \vskip -0.05in
    \caption{
        An overview of our proposed influence-guided coreset selection and evaluation pipeline for large-scale RNA sequence pretraining.
    }
    \label{fig: main}
    \end{center}
    \vskip -0.2in
\end{figure*}

Recent advances in biological foundation models (BioFMs) have enabled remarkable progress in tasks such as structure prediction, functional annotation, and molecular interaction modeling across RNA/DNA and protein sequences~\cite{chen2022rnafm, evo2, esm3, shen2024accurate, bytedance2025protenix}. 
Despite their success, these models typically rely on extremely large-scale pretraining data, which demand substantial computational, environmental, and reproducibility costs. 
For example, RNA-FM is trained on over 23 million RNA sequences~\cite{chen2022rnafm}, while ESM has scaled up to 2.78 billion protein sequences in its latest versions~\cite{esm3}, making it practically infeasible for most research groups to reproduce the full training process.

To promote open and sustainable development in BioFMs, we focus on investigating the potential of data pruning~\cite{phillips2017coresets, 2025coresetSurvey} as a means to reduce substantial computational overhead while maintaining competitive performance. 
Specifically, we explore whether a carefully selected coreset can be used to retrain from scratch with significantly fewer examples.
To the best of our knowledge, this direction has received little attention in the context of BioFMs pretraining.
While coreset selection has been studied in CV/NLP~\cite{2025coresetSurvey,diddee2024chasingrandom}, most existing approaches fall into two main categories: 
(i) those that rely on training dynamics~\cite{pleiss2020aum, DynamicUncertainty, DUAL};
(ii) those based on local density measures~\cite{DBLP:conf/nips/small2large, zhang2024tagcos}.
However, both types face fundamental limitations in the biological setting. 
First, the pretraining cost is prohibitively high, and most BioFMs do not publicly release training details, rendering training-dynamics-based methods inapplicable. 
Second, the millions-to-billions scale of biological sequences poses significant scalability barriers for methods that depend on pairwise similarity computations.

To address these limitations, we propose an influence-guided coreset selection framework that operates in a post-hoc manner, without requiring access to the full training process.
Specifically, our framework consists of two main stages: 
(i) estimating influence scores for individual training examples; 
(ii) selecting subsets based on these scores with tailored selection strategies.
First, grounded in the classical influence function framework~\cite{koh2017understanding}, we reformulate a scalable subset-based self-influence function that estimates the impact of each training example, replacing the need to compute Hessians over the entire training data.
To make this approximation theoretically sound, we introduce a key assumption—the subset-based ERM condition—which requires the model to be sufficiently trained on a small randomly sampled subset. When this condition is satisfied, the curvature around the subset can serve as a faithful surrogate for the full training curvature.
To further reduce the computational cost, we adopt a diagonal empirical Fisher matrix as a tractable curvature approximation, enabling scalable influence estimation even at the scale of biological foundation models.
Next, we introduce two influence-guided selection strategies designed to serve different objectives:
(1) Top-$k$ Influence-guided Selection (Top I);
(2) Coverage-centric Influence-guided Selection (CCI).
These two strategies allow us to explore how influential or diverse training examples contribute to representation learning.
To evaluate their effectiveness, we pretrain BioFMs on both RNA and protein sequences using only 0.2 million selected examples in each case and assess their performance across a comprehensive suite of downstream tasks.
An overview of our influence-guided coreset framework and evaluation pipeline is illustrated in \Cref{fig: main}, using RNA-FM as a representative example.
Our contributions are listed as follows:
\begin{itemize}[topsep=0.2pt, itemsep=0em, leftmargin=12pt, parsep=0.2pt]
    \item We propose a post-hoc influence-guided data pruning framework tailored for biological foundation models, eliminating the need for full training access.
    \item We provide a theoretical derivation of our reformulated influence function, which supports efficient approximation through curvature over randomly sampled subsets.
    \item We introduce two influence-guided coreset selection strategies: Top I and CCI, both of which help us to understand the representation ability of corresponding coresets.
    \item We first conduct extensive experiments on RNA-FM, demonstrating that our data pruning framework achieves competitive performance across multiple downstream tasks, and then validate its generalizability on ESM-C.
\end{itemize}

\section{Related Work}
\subsection{Data Pruning} 

Data pruning, also known as coreset selection, aims to identify a small yet representative subset of a large training corpus such that training on this subset yields performance comparable to using the full dataset~\cite{phillips2017coresets, 2025coresetSurvey}. 
One common approach assigns importance scores to individual training examples based on training dynamics—for example, EL2N~\cite{Coleman2020entropy}, AUM~\cite{pleiss2020aum}, and Dynamic Uncertainty~\cite{DynamicUncertainty}. 
These methods quantify informativeness from model prediction behaviors over the course of training and are efficient when the full training process is accessible. 
However, they require tracking model outputs across multiple epochs, which is often infeasible or prohibitively expensive in the context of BioFMs.
Another line of work focuses on local data density~\cite{DBLP:conf/nips/BeyondNeuralScalingLaws, DBLP:conf/nips/small2large, zhang2024tagcos}, where representativeness is measured by a sample’s proximity to its neighbors in the feature space. 
Yet such approaches typically rely on storing high-dimensional embeddings and performing pairwise similarity computations, which becomes computationally impractical for large-scale datasets.
An alternative line of research is grounded in influence functions~\cite{koh2017understanding}, which will be further discussed in \Cref{sec:prelimiaries}.
While theoretically grounded, influence functions require computing gradients and inverse Hessian, which becomes computationally intensive in large-scale settings.

\subsection{Biological Sequence Representation}

Learning effective representations of biological sequences—RNA, DNA, and proteins—is fundamental to a wide range of downstream tasks, including structure prediction, function annotation, and biomolecular interaction modeling~\cite{shen2024accurate}. Due to the intrinsic complexity of biological macromolecules, such as RNA’s hierarchical structure and protein folding patterns, traditional approaches often relied on hand-crafted features or shallow learning techniques specific to individual tasks.
Recent advances in biological foundation models (BioFMs)~\cite{chen2022rnafm, esm3, evo2} have demonstrated the effectiveness of large-scale self-supervised learning, particularly masked language modeling, in capturing intricate biological semantics directly from raw sequences. For RNA, models, such as RNA-FM~\cite{chen2022rnafm}, have achieved strong transfer performance across diverse tasks, including RNA type classification~\cite{amin2019evaluation}, CRISPR-Cas efficiency prediction~\cite{chuai2018deepcrispr}, and RNA-binding protein (RBP) interaction prediction~\cite{xu2023prismnet, zhu2023dynamic}. On the protein side, ESM~\cite{esm3} has shown promising results on protein function prediction~\cite{Flu}, protein structure prediction~\cite{netsurfp}, and interaction prediction~\cite{guo2008using}.
Detailed descriptions of RNA-FM and ESM models, including their architectures, are provided in Appendix~\ref{app: biofms}.
Despite their success, BioFMs suffer from high training costs and limited reproducibility, which further motivates the need for post-hoc data-efficient approaches, such as data pruning, that can identify informative training subsets without full retraining access.

\section{Methods}

\subsection{Preliminaries: Influence Function}\label{sec:prelimiaries}

Influence functions (IF) aim to understand the effect of individual training points on a model’s predictions,  
which can be instantiated as the task of estimating how the model’s output would change if a particular training point were removed~\cite{hampel1974influence,cook1980characterizations}.

Let $D_{\text{tr}} = \{ z_n = (x_n, y_n) \}_{n=1}^N$ be an i.i.d. training dataset.  
Empirical risk minimization (ERM) solves:
\[
\theta^\star := \arg\min_\theta \frac{1}{N} \sum_{n=1}^N \ell(z_n, \theta),
\]
where $\ell$ denotes a per-sample loss function and $\theta^\star$ is the resulting minimizer.
Next consider a validation set $ D_{\text{val}} = \{ z_m = (x_m, y_m) \}_{m=1}^M $.  
The influence of a training point $z_{\text{tr}} \in D_{\text{tr}}$ on a specific validation example $z_{\text{val}} \in D_{\text{val}}$ can be expressed as the excess loss: $\ell(z_{\text{val}}, \theta^\star_{z_{\text{tr}}}) - \ell(z_{\text{val}}, \theta^\star)$,
where $\theta^\star_{z_{\text{tr}}}$ is the solution to a perturbed ERM objective in which $z_{\text{tr}}$ is upweighted by a small amount $\epsilon$:
\[
    \theta^\star_{z_{\text{tr}}} := \arg\min_\theta \frac{1}{N} \sum_{n=1}^N \ell(z_n, \theta) + \epsilon \ell(z_{\text{tr}}, \theta).
\]
In particular, setting $\epsilon = -\frac{1}{N}$ corresponds to the removal of $z_{\text{tr}}$ from the training set.

Following \citet{koh2017understanding}, this excess loss can be approximated via a two-step procedure.

\paragraph{Step 1: Parameter change.}  
The shift in parameters due to perturbing $z_\text{tr}$ can be approximated by a Newton step:
\[
        \theta^\star_{z_\text{tr}} - \theta^\star \approx -\epsilon H_{\text{tr}}^{-1} g_{z_\text{tr}},
\]
where \( g_{z_{\text{tr}}} = \nabla_{\theta^\star} \ell(z_{\text{tr}}, \theta^\star) \) denotes the gradient of the loss w.r.t. $z_{\text{tr}}$, and \( H_{\text{tr}} = \frac{1}{N} \sum_{n=1}^N \nabla^2_{\theta^\star} \ell(z_n, \theta^\star) \) is the Hessian of the empirical loss over \( D_{\text{tr}} \), which captures the local curvature of the empirical risk around the training optimum~\cite{van2000asymptotic}.

\paragraph{Step 2: Loss change.}  
The change in the validation loss of the sample $z_\text{val}$ can then be estimated via first-order Taylor expansion:
\[
    \ell(z_\text{val}, \theta^\star_{z_\text{tr}}) - \ell(z_\text{val}, \theta^\star)
    \approx
    g_{z_\text{val}}^\top (\theta^\star_{z_\text{tr}} - \theta^\star),
\]
where $g_{z_\text{val}} = \nabla_{\theta^\star} \ell(z_\text{val}, \theta^\star)$.

\paragraph{Final form.}  
Combining the two steps yields the influence function of $z_\text{tr}$ to $z_\text{val}$ on $\theta^\star$:
\begin{equation}
\mathcal{I}(z_\text{tr}; z_\text{val}) := g_{z_\text{val}}^\top H_{\theta^\star}^{-1} g_{z_\text{tr}}.
\end{equation}

In addition, the influence function of $z_\text{tr}$ on the entire validation set $D_\text{val}$ can be extended as follows:
\begin{equation}
    \label{eq: standard influence function}
    \mathcal{I}(z_\text{tr}; D_\text{val})
    := \frac{1}{M} \sum_{m=1}^M g_{z_m}^\top H_\text{tr}^{-1} g_{z_\text{tr}},
\end{equation}
where $z_m \in D_\text{val}$ represents a validation sample.

By replacing $D_\text{val}$ with the training set $D_\text{tr}$ in \Cref{eq: standard influence function}, we obtain the self-influence function~\cite{koh2017understanding} of a training point:
\begin{equation}
    \label{eq: self influence}
    \mathcal{I}(z_\text{tr}; D_\text{tr})
    := \frac{1}{N} \sum_{n=1}^N g_{z_n}^\top H_\text{tr}^{-1} g_{z_\text{tr}},
\end{equation}
where $z_n \in D_\text{tr}$.
This allows us to quantify the effect of each training point in self-supervised settings. 
However, computing influence scores over large-scale training sets remains prohibitively expensive in both memory and computation.
To address this, we introduce scalable approximations in the following sections.

\subsection{Subset-Based Scalable Influence Estimation}

Direct computation of the standard self-influence function, as defined in Eq.~\ref{eq: self influence}, requires accessing the full training Hessian $H_{\text{tr}}$ and computing its inverse. For BioFMs like RNA-FM, which contain billions of parameters, this becomes infeasible in both memory and runtime.
Motivated by recent advances in curvature-based influence reformulations~\cite{ye2025robust}, we investigate whether influence can be approximated using curvature estimated over a small training subset.
To this end, we propose a two-step strategy:
(i) reformulate the influence function based on the training subset;
(ii) apply a lightweight inverse Hessian approximation to further reduce computational overhead.

\subsubsection{Reformulating Self-Influence via the Training Subset}

The classical self-influence function~\cite{koh2017understanding} assumes the model is trained to minimize empirical risk over the full training set, thereby allowing the influence of a single example to be approximated via the curvature of the training loss, i.e., $H_\text{tr}$.
To enable scalable estimation, we propose to replace $H_\text{tr}$ with $H_\text{sub}$ computed on a small random sampled training subset. 
To do so, we first assume that the model parameters are locally optimal with respect to this subset. This leads to the following assumption:
\begin{assumption}[Subset-based Empirical Risk Minimization]
\label{assumption:subset-erm}
Let $D_\text{sub} = \{ z_m \}_{m=1}^M \subset D_\text{tr}$ be a randomly sampled subset of the training set. We assume that the model parameters $\tilde{\theta}$ are obtained by minimizing the empirical risk over $D_\text{sub}$:
\[
\tilde{\theta} := \arg\min_\theta \frac{1}{M} \sum_{m=1}^{M} \ell(z_m, \theta).
\]
\end{assumption}

As described in \Cref{sec:prelimiaries}, we also decompose self-influence estimation into two steps:  parameter change and loss change.

\paragraph{Step 1: Parameter change.}  
Different from $\theta^\star_{z_\text{tr}}$ as described in \Cref{sec:prelimiaries}, we consider $\tilde{\theta}_{z_\text{tr}}$ defined as:
\[
    \tilde{\theta}_{z_\text{tr}}
    := \arg\min_\theta \frac{1}{M} \sum_{m=1}^M \ell(z_m, \theta)) + \epsilon \ell(z_\text{tr}, \theta).
\]
Therefore, the parameter change $\tilde{\theta}_{z_\text{tr}} - \tilde{\theta}$ can be approximated using a Newton step:
\begin{equation}
    \label{eq: parameter change on tilde theta}
    \tilde{\theta}_{z_\text{tr}} - \tilde{\theta}
    \approx
    - \epsilon \tilde{H}^{-1}_\text{sub} \tilde{g}_{z_\text{tr}},
\end{equation}
where $\tilde{g}_{z_\text{tr}} = \nabla_{\tilde{\theta}} \ell(z_\text{tr}, \tilde{\theta})$ is the gradient of the loss w.r.t. $\tilde{\theta}$ for training poit $z_\text{tr}$ and $\tilde{H}_{\text{sub}} = \frac{1}{M} \sum_{m=1}^M \nabla^2_{\tilde{\theta}} \ell (z_m, \tilde{\theta})$ is the Hessian matrix computed over $D_\text{sub}$.

\paragraph{Step 2: Loss change.}  
Previous studies~\cite{DBLP:conf/nips/gex} have shown that second-order approximations yield more accurate estimates of loss changes.
Therefore, we adopt a second-order Taylor expansion to estimate the loss change:
\begin{equation}
\label{eq: risk change on tilde theta}
\begin{aligned}
& \ell(z_{\text{sub}}, \tilde{\theta}_{z_{\text{tr}}}) - \ell(z_{\text{sub}}, \tilde{\theta}) \\
\approx\ ~& \tilde{g}_{z_{\text{sub}}}^\top \Delta\theta 
+ \frac{1}{2} \Delta\theta^\top \nabla^2_{\tilde{\theta}} \ell(z_{\text{sub}}, \tilde{\theta}) \Delta\theta,
\end{aligned}
\end{equation}
where $\Delta \theta = \tilde{\theta}_{z_\text{tr}} - \tilde{\theta}$ and $\tilde{g}_{z_\text{sub}} = \nabla_{\theta} \ell(z_\text{sub}, \tilde{\theta})$.

\paragraph{Final form.}
Combining Eq.~\ref{eq: parameter change on tilde theta} and Eq.~\ref{eq: risk change on tilde theta}, we can have self-influence function of $z_\text{tr}$ to $z_\text{sub}$ on $\tilde{\theta}$:
\begin{equation}
\label{eq: IF on tilde theta}
\begin{aligned}
\mathcal{I}(z_\text{tr}, z_\text{sub}&) 
:= \tilde{g}_{z_\text{sub}}^\top \tilde{H}^{-1}_\text{sub} \tilde{g}_{z_\text{tr}} \\
& + \frac{1}{2} \epsilon \tilde{g}_{z_\text{tr}}^\top \tilde{H}^{-1}_\text{sub} \nabla^2_{\tilde{\theta}} \ell(z_{\text{sub}}, \tilde{\theta}) \tilde{H}^{-1}_\text{sub} \tilde{g}_{z_\text{tr}}.
\end{aligned}
\end{equation}
Under Assumption~\ref{assumption:subset-erm}, i.e., $\tilde{g}_\text{sub} \to 0$, we will directly drop $\tilde{g}_{z_\text{sub}}^\top \tilde{H}^{-1}_\text{sub} \tilde{g}_{z_\text{tr}}$ later.
By extending Eq.~\ref{eq: IF on tilde theta}, we can measure the self-influence of $z_\text{tr}$ on $D_\text{sub}$:
\begin{equation}
\label{eq: sub_IF}
    \mathcal{I}(z_\text{tr}, D_\text{sub})
    := \tilde{g}_{z_\text{tr}}^\top \tilde{H}^{-1}_\text{sub} \tilde{g}_{z_\text{tr}}.
\end{equation}

Given the uniform training objective and the flat loss landscape observed in large models~\cite{chen2025llmflat}, the curvature over a random subset $D_\text{sub}$ is expected to approximate that of the full training set $D_\text{tr}$.
Therefore, we have the following approximation result:

\begin{proposition}[Subset-based Self-Influence Approximation]
\label{prop:subset-self}
Under Assumption~\ref{assumption:subset-erm}, and assuming the loss landscape around $\tilde{\theta}$ is flat, the self-influence of $z_\text{tr}$ on $D_\text{tr}$ can be approximated by
\(\mathcal{I}(z_\text{tr}, D_\text{sub})\)
\end{proposition}

The derivation and theoretical discussion are provided in \Cref{appendix: derivation}.

\subsubsection{Efficient Approximation for Inverse Hessian} 
\label{sec: diagonal fisher}

Despite the reformulation in Eq.~\ref{eq: sub_IF}, computing the full Hessian inverse $H^{-1}_\text{sub}$ remains computationally intractable. The total complexity amounts to $O(M \cdot d^2 + d^3)$, where $M$ is the number of subset examples and $d$ is the number of model parameters. This cost is prohibitive for large-scale BioFMs with billions of parameters.

To address this issue, we note that BioFMs are typically trained with a negative log-likelihood loss and, under Assumption~\ref{assumption:subset-erm}, the model is well-trained on the subset, allowing us to approximate the Hessian $H_\text{sub}$ using the empirical Fisher information matrix~\cite{DBLP:journals/corr/RevisitingNaturalGradient}.
Specifically, the empirical Fisher matrix over the training subset is given by:
\begin{equation}
\label{eq: empirical fisher}
\tilde{F}_{\text{sub}} := \frac{1}{M} \sum_{m=1}^M \tilde{g}_{z_m} \tilde{g}_{z_m}^\top,
\end{equation}
where \( \tilde{g}_{z_m} := \nabla_{\tilde{\theta}} \ell(z_m, \tilde{\theta}) \) denotes the gradient of the loss w.r.t. the pretrained model parameters \( \tilde{\theta} \) on the subset sample \( z_m \in D_{\text{sub}} \).

To further reduce computational cost, we follow the practice commonly adopted in conjugate gradient methods~\cite{DBLP:conf/nips/Topmoumoute,DBLP:conf/icml/Nomorepesky,DBLP:conf/icml/OptimizingNeural} and adaptive optimizers such as Adam~\cite{DBLP:journals/corr/adam}, where the curvature matrix is approximated by its diagonal.  
In this spirit, we apply a diagonal approximation to the empirical Fisher matrix, which yields:
\begin{equation}
\label{eq: diag fisher}
    \text{diag}(\tilde{F}_{\text{val}}) := \frac{1}{M} \sum_{m=1}^M \tilde{g}_{z_m} \odot \tilde{g}_{z_m},
\end{equation}
where \( \odot \) denotes the element-wise (Hadamard) product. 
Therefore, the inverse Hessian can be approximated as:
\begin{equation}
\label{eq: fisher inverse approx}
    \tilde{H}_{\text{val}}^{-1} \approx \text{diag}(\tilde{F}_{\text{val}})^{-1}.
\end{equation}
Substituting this into Eq.~\ref{eq: sub_IF}, we obtain the final scalable approximation for subset-based self-influence function:
\begin{equation}
\label{eq: scalable IF}
    \mathcal{I}(z_\text{tr}, D_\text{sub}) := \tilde{g}_{z_\text{tr}}^\top \text{diag}(\tilde{F}_{\text{sub}})^{-1} \tilde{g}_{z_\text{tr}}.
\end{equation}
This approximation enables influence estimation with linear complexity, reducing the overall computational cost from $O(M \cdot d^2 + d^3)$ to $O(M \cdot d)$—making it practical for large-scale BioFMs with billions of parameters.

\begin{remark}
\label{remark:subset-erm}
The effectiveness of both the subset-based influence approximation and the Fisher-based curvature estimation hinges on Assumption~\ref{assumption:subset-erm}, which assumes that the model is well-trained on the selected subset. In practice, we find that a light-weight fine-tuning (e.g., one epoch) on the subset is sufficient to satisfy this condition. The associated cost is negligible compared to the overall influence estimation pipeline, making the approach practical for large-scale BioFMs.
\end{remark}

\subsection{Influence-guided Coreset Selection Strategy}
\label{sec: ccs}

Building on the theoretical properties of influence functions~\cite{koh2017understanding}, we propose two selection strategies: Top-$k$ Influence-guided Selection and Coverage-centric Influence-guided Selection, tailored for coreset construction.

\paragraph{Top-$k$ Influence-guided Selection.}
Since the influence score of a training example quantifies its estimated contribution to the model’s performance, selecting the top-$k$ examples with the highest influence naturally prioritizes those with the greatest potential to affect generalization. 
This simple yet principled approach aligns with the coreset selection goal of retaining the most informative points.

\paragraph{Coverage-centric Influence-guided Selection.}
Recent findings~\cite{DBLP:conf/nips/BeyondNeuralScalingLaws,DBLP:conf/iclr/moderate} have revealed that the utility of different examples depends on the available data regime. 
Specifically, when only a small amount of data is retained, it is often more effective to preserve the \emph{easiest} examples, as they convey coarse-grained information about the target function and help avoid overfitting. 
In contrast, hard examples typically provide fine-grained information, which becomes useful only when the model has already captured the basics of the distribution. 
Under extreme pruning, focusing solely on the hardest or most influential examples may hinder learning, as outliers or rare cases may dominate the subset while the underlying structure of the data remains underrepresented~\cite{DBLP:conf/emnlp/DatasetCartography}.
Motivated by this, similar to \citet{DBLP:conf/iclr/ccs}, we apply stratified sampling over the influence score distribution, which ensures both easy and hard examples remain under extreme pruning. For full algorithmic details, please refer to \Cref{app:ccs-algorithm}.

\paragraph{Comparison.}
Top-$k$ influence-guided selection emphasizes informativeness and parameter sensitivity, whereas Coverage-centric Influence-guided selection focuses on representational diversity and robustness under high pruning rates.
To further explore these two strategies, we empirically evaluate both strategies in \Cref{sec:exp}, and discuss when each approach may be preferable.

\section{Experiments}
\label{sec:exp}

To validate the effectiveness and generalizability of our post-hoc influence-guided data pruning framework, we conduct experiments on both RNA and protein foundation models, namely RNA-FM~\cite{chen2022rnafm} and ESM-C~\cite{esm3}.

Considering the prohibitive cost of pretraining BioFMs\footnote{RNA-FM was trained on 23 million sequences using 8 A100 GPUs over 30 days~\cite{chen2022rnafm}.}, we adopt an extreme data pruning evaluation setting, where only 0.2 million sequences are retained and used for pretraining.
For RNA-FM, we conduct data pruning over the entire 23M-sequence training data, i.e., over 99\% data pruning. 
In contrast, given the inaccessibility of the full 2.78-billion protein sequence used for ESM-C pretraining, we instead collect around 4.5 million protein sequences from UniRef50~\cite{suzek2007uniref} and conduct data pruning over this.

After data pruning via different selection strategies, we pretrain BioFMs on 0.2 million sequences from scratch for 10 epochs and evaluate them across a range of downstream tasks.
Further training details are provided in \Cref{app: Implementation Details}.

\subsection{Selection Strategies and Baselines} 
We consider the following selection strategies and baselines in our experiments:

\begin{itemize}
\item \textbf{RNA-FM / ESM-C}: The original foundation model trained on the full dataset.

\item \textbf{Raw}: Untrained model.  

\item \textbf{Random}: Uniform random sampling of 0.2M sequences (matching our pruning budget) or 2M sequences.

\item \textbf{Top I}: Applies Top-$k$ influence-guided selection with our subset-based self-influence. 

\item \textbf{CCI}: Applies coverage-centric influence-guided selection with our subset-based self-influence.
\end{itemize}

\noindent \textbf{Reamark.} ~
For Top I and CCI, we consider two variants: the default version performs a lightweight fine-tuning step on a randomly selected 0.2-million training subset to adapt influence estimation (see Remark~\ref{remark:subset-erm}), 
whereas the (w/o ft) variant uses scores computed from the initial pretrained model without any adaptation.
\subsection{Downstream Tasks and Evaluation Metrics}

To comprehensively evaluate model performance, we assess RNA-FM and ESM-C variants pretrained on different coresets, as well as full-data and raw baselines raw and full-data baselines, across a diverse suite of downstream RNA and protein understanding tasks.

\begin{table*}[t]
\center
\footnotesize
\begin{tabular}{@{}l@{\hskip 8.5pt}c@{\hskip 8.5pt}c@{\hskip 8.5pt}c@{\hskip 8.5pt}c@{\hskip 8.5pt}c@{\hskip 8.5pt}c@{}}
\toprule
\multirow{2}{*}{Methods} & \multirow{2}{*}{Data Size} & \multicolumn{2}{c}{TypeCls} & Modif & \multicolumn{2}{c}{CRI-On}\\
\cmidrule(lr){3-4} \cmidrule(lr){5-5} \cmidrule(lr){6-7}
 & & ACC(\%) & F1(\%) & AUC(\%) & SC(\%) & MSE~$\downarrow$\\
\midrule
RNA-FM  & 23M & 91.93 & 91.87 & 94.98 &31.87 & .0118\\
\midrule
Raw & 0M & 79.46 & 78.96 & 90.71 & 22.48 & .0261\\
Random & 2M & 82.21 & 82.01 & 92.82 & 26.72 & .0158\\
\midrule
Random  & 0.2M & 82.15 & 81.97 & 91.86 & 26.67 & .0161\\
Top I (w/o ft) &  0.2M & 81.07 & 81.21 & 92.94 & \underline{28.60} & .0151\\
CCI (w/o ft) & 0.2M & 80.60 & 80.37 & \underline{93.31} & 26.96 & .0150\\
Top I & 0.2M & \underline{82.51} & \underline{82.53} &93.20 & 27.08 & \underline{.0149}\\
CCI & 0.2M & \textbf{82.88} & \textbf{83.12} & \textbf{93.86} & \textbf{32.90} & \textbf{.0135}\\
\bottomrule
\end{tabular}
\vspace{-0.1em}
\caption{Performance of different coresets across three function and engineering prediction tasks (RNA Type Classification, RNA Modification Prediction, and CRISPR On-Target Prediction). \textbf{Bold} denotes the best results and \underline{underline} denotes the second-best results.}
\vspace{-0.1em}
\label{tab: main results}
\end{table*}

\subsubsection{RNA-FM}
The downstream tasks~\cite{chen2022rnafm, DBLP:conf/nips/Beacon} can be categorized as follows:

\begin{itemize}
\item \textbf{Function Prediction}: 
(1) RNA Type Classification (TypeCls)~\cite{amin2019evaluation}, evaluated using accuracy and F1 score;
(2) RNA Modification Prediction (Modif)~\cite{duan2019dynamic}, evaluated using AUC.

\item \textbf{Engineering Prediction}:
CRISPR On-Target Prediction (CRI-On)~\cite{chuai2018deepcrispr}, evaluated using SC and MSE.

\item \textbf{Structure Prediction}: 
(1) Secondary Structure Prediction on bpRNA~\cite{danaee2018bprna}, evaluated using precision, recall, F1 score, and MCC;  
(2) Distance Map Prediction~\cite{chen2022rnafm}, evaluated using R$^2$, Spearman correlation (SC), MAE, and MSE;
(3) Contact Map Prediction~\cite{chen2022rnafm}, evaluated using Top-$L$ precision.

\item \textbf{Interaction Prediction}:
RBP-RNA Interaction prediction~\cite{chen2022rnafm}, evaluated using accuracy, AUC, and AUPR.
\end{itemize}

\subsubsection{ESM}
The downstream tasks~\cite{xu2022peer} can be categorized as follows:

\begin{itemize}
\item \textbf{Localization Prediction}: Binary Localization Prediction (Bin)~\cite{almagro2017deeploc}, evaluated using accuracy.

\item \textbf{Structure Prediction}: Secondary Structure Prediction (SS)~\cite{netsurfp}, evaluated using accuracy.

\item \textbf{Interaction Prediction}: PPI Affinity Prediction(Aff)~\cite{moal2012skempi}, evaluated using MAE and RMSE.

\end{itemize}

Further fine-tuning details are provided in \Cref{app: experiment details}.

\section{Results and Analysis}
\label{sec: results}

In this section, we first conduct a comprehensive evaluation of our influence-guided data pruning framework on RNA-FM, covering a broad range of RNA-specific tasks.
To assess its generalizability, we further apply the same framework to ESM-C, a protein foundation model, demonstrating its robustness across distinct biomolecular modalities.

\subsection{Results on RNA-FM}
\label{sec: model-rna-fm}

\subsubsection{Overall Performance across RNA Tasks}

We mainly categorize RNA downstream tasks into two groups: Function and Engineering Prediction and Structure and Interaction Prediction.

\medskip

\noindent \textbullet~\textbf{Function and Engineering Prediction}
We first examine RNA Type Classification (TypeCls), Modification Prediction (Modif), and CRISPR On-Target Prediction (CRI-On). The complete results are reported in Table~\ref{tab: main results}.
Across all three tasks, both Top I and CCI consistently outperform the Random baseline.
Although the performance margins of TypeCls over Random may appear modest, achieving consistent improvements across all tasks still demonstrates the effectiveness of our influence-guided approach and highlights the value of exploring informed pruning for BioFMs.
Notably, CCI consistently achieves the best performance across all three tasks.
Its superior results suggest that incorporating coverage and diversity patterns is more beneficial for the model to capture functional patterns of RNA sequences during self-supervised pretraining.
Furthermore, on CRISPR On-Target Prediction, CCI even surpasses the full RNA-FM model trained on all 23 million sequences, which demonstrates that, in certain scenarios, high-quality, task-relevant information can be effectively preserved within a drastically reduced training subset.

\medskip

\noindent \textbullet~\textbf{Structure and Interaction Prediction}
We then examine Secondary Structure Prediction on bpRNA, Distance Map Prediction, Contact Map Prediction, and RBP-RNA Interaction Prediction.
The complete results are reported in Table~\ref{tab:all-tasks}.
Both Top I and CCI again surpass the Random baseline across nearly all metrics, reinforcing the effectiveness of our influence-guided data pruning approach.
Interestingly, Top I demonstrates notable advantages in structure- and interaction-related tasks, both of which are highly dependent on the underlying RNA structural properties.
In particular, in Contact Map Prediction (\Cref{tab:rnact}), Top I even surpasses RNA-FM on both Top-1.0L and Top-0.5L precision metrics.
This suggests that, within RNA datasets, examples with higher self-influence scores tend to encode richer structural information, which can be effectively leveraged by models during self-supervised pretraining.
Moreover, across all four tasks in this category, the performance gap between Top I and full-data RNA-FM remains remarkably small.
This indicates that, for structure- and interaction-related tasks, a compact subset consisting of the most self-influential samples can effectively replace the full 23-million-sequence dataset for RNA-FM pretraining.

\subsubsection{Data Redundancy in RNA Training Data}

From the results reported in Table~\ref{tab: main results}, we observe that both Top I and CCI consistently outperform the Random 2M baseline across all function and engineering prediction tasks, while using only 10\% of the data volume (0.2M vs. 2M sequences).
This performance gain, achieved under a significantly smaller data budget, provides strong evidence that the RNA training data contains considerable redundancy.
Such empirical evidence highlights the potential and necessity of exploring data pruning or coreset selection techniques tailored to RNA pretraining, especially under large-scale pretraining scenarios.

\subsubsection{Ablation: Necessity of Adaptation for Subset-based Influence}

To validate the necessity of fine-tuning on the subset prior to influence estimation, we compare our influence-guided selection strategies with and without adaptation (denoted as w/o ft) in both Table~\ref{tab: main results} and Table~\ref{tab:all-tasks}.
In all cases, the adapted variants (Top I and CCI) consistently outperform their non-adapted counterparts, demonstrating the importance of aligning the model with Assumption~\ref{assumption:subset-erm} before computing influence.
These results confirm that the lightweight adaptation step described in Remark~\ref{remark:subset-erm} is critical for reducing estimation error in both influence scores and curvature approximations.
Properly aligning the model to the subset enables us to obtain more reliable self-influence estimates, thus facilitating more effective and principled data pruning.

\begin{table*}[t]
\centering
{
\footnotesize

\begin{subtable}[t]{0.48\textwidth}
\centering
\begin{tabular}{@{}l@{\hskip 8.5pt}c@{\hskip 8.5pt}c@{\hskip 8.5pt}c@{\hskip 8.5pt}c@{\hskip 8.5pt}c@{}}
\toprule
Methods & Data Size & Pre(\%) & Rec(\%) & F1(\%) & MCC(\%) \\
\midrule
RNA-FM & 23M & 66.14 & 62.24 & 62.20 & 63.01 \\
\midrule
Random   & 0.2M & 59.75 & 55.59 & 55.60 & 56.49 \\
Top I (w/o ft) & 0.2M & 59.74 & \underline{58.22} & \underline{56.95} & \underline{57.76} \\
CCI (w/o ft)  & 0.2M & 59.30 & 57.20 & 56.10 & 57.00 \\
Top I & 0.2M & \underline{59.76} & \textbf{58.27} & \textbf{57.05} & \textbf{57.85} \\
CCI  & 0.2M & \textbf{60.29} & 56.33 & 56.36 & 57.14 \\
\bottomrule
\end{tabular}
\caption{Secondary Structure Prediction on bpRNA.}
\label{tab:ssp}
\end{subtable}
\hfill
\begin{subtable}[t]{0.48\textwidth}
\centering
\begin{tabular}{@{}l@{\hskip 8.5pt}c@{\hskip 8.5pt}c@{\hskip 8.5pt}c@{\hskip 8.5pt}c@{\hskip 8.5pt}c@{}}
\toprule
Methods & Data Size & R$^2$(\%) & SC(\%) & MAE~$\downarrow$ & MSE~$\downarrow$ \\
\midrule
RNA-FM  & 23M & 83.26 & 89.21 & .5665 & .6650\\
\midrule
Random & 0.2M & 76.71 & 84.90 & .7176 & 1.037\\
Top I (w/o ft) & 0.2M & 75.91 & 84.13 & .7284 & 1.057\\
CCI (w/o ft) & 0.2M & 76.80 & 84.95 & .7254 & 1.045\\
Top I & 0.2M & \textbf{79.25} & \textbf{86.47} & \textbf{.6745} & \textbf{.9215}\\
CCI & 0.2M & \underline{77.98} & \underline{85.59} & \underline{.6937} & \underline{.9861}\\
\bottomrule
\end{tabular}
\caption{RNA Distance Map Prediction.}
\label{tab:rnadist}
\end{subtable}

\vspace{1em}

\begin{subtable}[t]{0.48\textwidth}
\centering
\begin{tabular}{@{}l@{\hskip 8.5pt}c@{\hskip 8.5pt}c@{\hskip 8.5pt}c@{\hskip 8.5pt}c@{\hskip 8.5pt}c@{}}
\toprule
\multirow{2}{*}{Methods} & \multirow{2}{*}{Data Size} & \multicolumn{4}{c}{Long-Range Top Precision~(\%)}\\
\cmidrule(lr){3-6}
 & & L:1.0L & L:0.5L & L:0.2L & L:0.1L\\
\midrule
RNA-FM  & 23M & 93.93 & 98.28 & 99.62 & 99.86\\
\midrule
Random & 0.2M & 94.18 & 98.20 & 99.28 & 99.31\\
Top I (w/o ft) & 0.2M & 93.94 & 98.05 & 99.06 & 98.99\\
CCI (w/o ft) & 0.2M & 93.86 & 98.22 & \underline{99.32} & \textbf{99.46}\\
Top I & 0.2M & \textbf{94.36} & \textbf{98.41} & \textbf{99.39} & \underline{99.39}\\
CCI & 0.2M & \underline{94.20} & \underline{98.26} & 99.14 & 99.21\\
\bottomrule
\end{tabular}
\caption{RNA Contact Map Prediction.}
\label{tab:rnact}
\end{subtable}
\hfill
\begin{subtable}[t]{0.48\textwidth}
\centering
\begin{tabular}{@{}l@{\hskip 8.5pt}c@{\hskip 8.5pt}c@{\hskip 8.5pt}c@{\hskip 8.5pt}c@{}}
\toprule
Methods & Data Size & ACC(\%) & AUPR(\%) & AUC(\%) \\
\midrule
RNA-FM  & 23M & 72.47 & 67.19 & 79.68\\
\midrule
Random & 0.2M & 69.65 & 62.14 & 75.97\\
Top I (w/o ft) & 0.2M & \underline{70.62} & \underline{63.40} & \textbf{77.16}\\
CCI (w/o ft) & 0.2M & 69.10 & 61.33 & 75.45 \\
Top I & 0.2M & \textbf{71.25} & \textbf{63.63} & \underline{76.97} \\
CCI & 0.2M & 69.46 & 62.10 & 76.04 \\
\bottomrule
\end{tabular}
\caption{RBP–RNA Interaction Prediction.}
\label{tab:rbp}
\end{subtable}

}
\vspace{-0.3em}
\caption{Performance comparison of different coreset selection strategies across four structure and interaction prediction tasks for RNA understanding. \textbf{Bold} denotes the best results and \underline{underline} denotes the second-best results.}
\label{tab:all-tasks}
\vspace{-0.5em}
\end{table*}

\begin{table}[t]
\begin{center}
{
\footnotesize
\begin{tabular}{@{}l@{\hskip 8.5pt}c@{\hskip 8.5pt}c@{\hskip 8.5pt}c@{\hskip 8.5pt}c@{\hskip 8.5pt}c@{}}
\toprule
\multirow{2}{*}{Methods} & \multirow{2}{*}{Data Size} & Bin & SS & \multicolumn{2}{c}{Aff}\\
\cmidrule(lr){3-3} \cmidrule(lr){4-4} \cmidrule(lr){5-6}
 & & ACC(\%) & ACC(\%) & MAE~$\downarrow$ & RMSE~$\downarrow$\\
\midrule
ESM-C  & 2.78B & 91.63 & 86.10 & 1.92 & 2.44 \\
\midrule
Random & 2M & 75.76 & 67.20 & 2.39 & 2.87 \\
\midrule
Random  & 0.2M & 73.64 & 66.18 & 2.51 & 3.01 \\
Top I & 0.2M & \underline{77.13} & \underline{69.34} & \textbf{2.06} & \textbf{2.64} \\
CCI & 0.2M & \textbf{79.25} & \textbf{71.48} & \underline{2.14} & \underline{2.69}\\
\bottomrule
\end{tabular}
}
\end{center}
\vspace{-0.5em}
\caption{Performance of different coresets across three different downstream prediction tasks (Binary Localization Prediction, Secondary Structure Prediction, and PPI Affinity Prediction). \textbf{Bold} denotes the best results and \underline{underline} denotes the second-best results.}
\vspace{-0.3em}
\label{tab: protein results}
\end{table}

\subsection{Results on ESM}
\label{sec: model-esm}

To assess the generalizability of our influence-guided data pruning framework, we apply it to the protein foundation model ESM-C.
We evaluate model performance on three representative downstream tasks: Binary Localization Prediction (Bin), Secondary Structure Prediction (SS), and Protein–Protein Interaction Affinity Prediction (Aff).
The complete results are displayed in \Cref{tab: protein results}.

Our influence-guided data pruning strategies—Top I and CCI—consistently outperform both Random baselines (with 0.2M and 2M samples) across all three tasks, thereby verifying the effectiveness of our pruning framework in the protein domain.
Moreover, this observation echoes our results on RNA-FM and suggests that the protein sequence dataset also exhibits a high level of redundancy, further underscoring the potential of data pruning.

Although the models pretrained on Top I or CCI coresets still exhibit a performance gap compared to the original ESM-C, this discrepancy can be attributed to the substantial gap in data scale (0.2M vs. 2.78B).
Due to current resource constraints, we leave the exploration of more suitable coreset sizes for protein foundation models as future work.

Nevertheless, these results still provide encouraging evidence that influence-guided data pruning holds promise across both RNA and protein domains, even under extremely limited data budgets.

\section{Conclusion}

In this work, we investigate the problem of data pruning for pretraining biological foundation models (BioFMs) at scale, aiming to alleviate the substantial computational demands posed by large-scale pretraining.
To this end, we introduce a post-hoc influence-guided data pruning framework that incorporates two complementary selection strategies—Top-$k$ Influence (Top I) and Coverage-Centric Influence (CCI)—to enable scalable and effective coreset construction.
Our experiments demonstrate that the proposed framework consistently outperforms random selection across both RNA and protein domains, while in RNA structure prediction tasks, it even achieves performance comparable to the original RNA-FM using less than 1\% of the full 23-million-sequence training set. This demonstrates the effectiveness of our data pruning framework.
Furthermore, models trained on the selected coresets even surpass counterparts trained on ten times more data via random sampling, revealing significant redundancy in current biological pretraining datasets, which underscores the potential of data pruning in BioFM pretraining.
Looking forward, our work offers a promising pathway toward training high-performing BioFMs on compact yet informative subsets, which can facilitate more reproducible, accessible, and sustainable biological AI research.

\section*{Acknowledgments}
This work was supported by Shenzhen Medical Research Fund (Grant No. A2503002 to Yu Li), the Major Project of Guangzhou National Laboratory
(Grant No. GZNL2024A01003, GZNL2023A02007, GZNL2025C02028 to Jiao Yuan), the National Natural Science Foundation of China (Grant No. 32400547 to Jiao Yuan and Grant No. 62472097 to Weizhong Zhang), High-Quality Development Project of Shanghai Municipal Commission of Economy and Informatization (Grant No. 2024-GZL-RGZN-02010 to Weizhong Zhang), AI for Science Foundation of Fudan University (FudanX24AI028 to Weizhong Zhang), Pearl River Talent Recruitment Program (2023QN10Y296 to Jiao Yuan), Guangzhou Young Top Talent Program, National Key R\&D Program of China (2023YFF1204701 to Jiao Yuan), the Chinese University of Hong Kong (CUHK; award numbers 4937025, 4937026, 5501517 and 5501329 to Yu Li), the IdeaBooster Fund (IDBF23ENG05 and IDBF24ENG06 to Yu Li), partially supported by a grant from the Research Grants Council of the Hong Kong Special Administrative Region (Hong Kong SAR), China (project no. CUHK 24204023 to Yu Li), a grant from the Innovation and Technology Commission of the Hong Kong SAR, China (project no. GHP/065/21SZ and ITS/247/23FP to Yu Li), and the Research Matching Grant Scheme at CUHK (award numbers 8601603 and 8601663 to Yu Li) from the Research Grants Council, Hong Kong SAR, China.

\bibliography{main}

\clearpage
\appendix
\onecolumn

\setcounter{secnumdepth}{2}

\setlength{\parindent}{0pt}

\begin{center}
{\LARGE\bf Technical Appendix to ``Investigating Data Pruning for Pretraining Biological Foundation Models at Scale''}
\end{center}


\section{Implementation Details}
\label{app: Implementation Details}

\subsection{Coreset Selection Algorithm}
\label{app:ccs-algorithm}

We first sort the dataset by influence score and prune a fixed proportion $\beta$ of the most influential (i.e., hardest) examples.
We then partition the remaining examples into $k$ non-overlapping strata, where each stratum spans an equal-width range of influence scores. 
Within each stratum, we allocate a fixed portion of the total sampling budget. 
If a stratum contains fewer examples than its budget, the remaining budget is evenly redistributed among the other strata. 
For details, we provide the full pseudocode in \Cref{alg:coreset}. Specifically, we set the pruning rate $\alpha=90$, the hard cutoff rate $\beta=5$, and the number of strata $k=50$ for our experiments.

\begin{algorithm}[ht]
    \small
    \caption{Coverage-centric Influence-based Selection (CCI)}\label{alg:coreset}
    \begin{algorithmic}[1]
    \Require$D=\{(z_i, \mathcal{I}_i)\}_{i=1}^n$: dataset with the influence score for each example; 
    $\alpha$: dataset pruning rate; $\beta$: hard cutoff rate ($\beta \leq 1 - \alpha$); $k$: the number of strata. 
    \vspace{0.4em}
    \hrule
    \vspace{0.4em}
    \State $D' \gets D \setminus \{\lfloor n * \beta \rfloor$ hardest examples $\}$ \Comment{Prune hard examples first}

    \State $R_1, R_2, ..., R_k \gets \text{Split scores in $D'$ into $k$ ranges with an even range width}$\;
    
    \State $\mathcal{B} \gets \{\mathbb{B}_i,: \mathbb{B}_i \text{ consists of examples whose scores are in } R_i, i = 1...k\}$;

    \State $m \gets n \times (1-\alpha)$ \Comment{m is total budget across all strata}
    \State $D_c \gets \varnothing$ \Comment{Initialize the coreset}
    \While{$\mathcal{B} \neq \varnothing$}
        \State $\displaystyle \mathbb{B}_{min} \gets \argmin_{\mathbb{B} \in \mathcal{B} |\mathbb{B}|}$ \Comment{Select the stratum with fewest examples}
        \State $m_B \gets \min\{|\mathbb{B}_{min}|, \lfloor \frac{m}{|\mathcal{B}|}\rfloor\}$  \Comment{Compute budget for selected stratum}
        \State $D_B \gets$ randomly sample $m_B$ examples from $\mathbb{B}_{min}$ \;
        \State $D_c \gets D_c \cup D_B$ \Comment{Update the coreset}
        \State $\mathcal{B} \gets \mathcal{B} \setminus \{\mathbb{B}_{min}\}$ \Comment{Done with selected straum}
        \State $m \gets m - m_B$ \Comment{Update total budget for remaining strata}
    \EndWhile
    \State \textbf{return} $D_c$ \Comment{Return the final coreset}
    \end{algorithmic}
\end{algorithm}

\subsection{Model Details of Biological Foundation Models}
\label{app: biofms}

\paragraph{RNA-FM}(RNA Foundation Model)~\cite{chen2022rnafm} is a self-supervised deep learning framework based on the BERT architecture\cite{devlin2018bert}, employing 12 bidirectional Transformer encoder layers to extract biologically meaningful representations from RNA sequences. It is pretrained on 23 million non-coding RNA sequences using a masked language modeling (MLM) objective.
Following pretraining, RNA-FM can be fine-tuned for various downstream tasks, including secondary structure prediction~\cite{danaee2018bprna}, contact map prediction~\cite{leontis2012nonredundant}, RBP interaction modeling~\cite{xu2023prismnet, zhu2023dynamic}, RNA type classification~\cite{amin2019evaluation}, and CRISPR-Cas efficiency prediction~\cite{chuai2018deepcrispr}.

\paragraph{ESM} (Evolutionary Scale Modeling)~\cite{esm3} is a family of large-scale protein language models developed by Meta, trained on protein sequences using masked language modeling (MLM). While early ESM versions adopted transformer architectures of varying sizes, the latest release includes ESM-3 and ESM-C, both pretrained on 2.78 billion protein sequences and equipped with architectural advancements, i.e., flash attention~\cite{dao2022flashattention}.
ESM models have been widely applied to diverse downstream tasks, including protein fluorescence prediction\cite{Flu}, stability estimation~\cite{rocklin2017global}, protein contact map prediction~\cite{alquraishi2019proteinnet}, and protein–protein interaction (PPI) prediction~\cite{guo2008using}.
Among these, ESM-C is a compact (600M parameters) variant specifically optimized for efficient embedding extraction, offering a favorable trade-off between performance and computational cost.
In this work, we build our coreset selection study upon ESM-C, given its demonstrated effectiveness and practical efficiency in large-scale representation learning.

\subsection{Training RNA-FM on Coresets}

After selecting a 0.2M-sequence coreset, we retrain the RNA-FM model from scratch using this compact dataset. Training is conducted on 2 NVIDIA A100 GPUs for 10 epochs with a batch size of 16.

We use the AdamW optimizer with a base learning rate of $1 \times 10^{-4}$ and a weight decay of 0.01. The learning rate follows an inverse square root decay schedule with a linear warmup phase over the first 2{,}000 steps.

\subsection{Training ESM-C on Coresets}

After selecting a 0.2M-sequence coreset, we retrain the ESM-C model from scratch using this compact dataset. Training is conducted on 2 NVIDIA A100 GPUs for 10 epochs with a batch size of 16.

We use the AdamW optimizer with a base learning rate of $1 \times 10^{-4}$ and a weight decay of 0.01. The learning rate follows an inverse square root decay schedule with a linear warmup phase over the first 2{,}000 steps.

\subsection{Post-hoc Subset Fine-tuning}
\label{app:vm-finetune}

To better satisfy Assumption~\ref{remark:subset-erm} discussed in Remark~\ref{remark:subset-erm}, we perform one additional epoch of masked language model (MLM) fine-tuning on the randomly selected 0.2M-sequence training subset with a batch size of 16 before computing influence scores. 

The MLM objective follows standard settings: we randomly mask 15\% of tokens in each input sequence. Optimization is performed using the AdamW optimizer with a learning rate of $1 \times 10^{-5}$, weight decay of $0.01$, and a linear warmup schedule over 50 steps. We keep all model parameters trainable and fine-tune the entire RNA-FM model for 1 epoch.

\subsection{Further details}

Throughout the data pruning process, all random seeds are fixed to 0 for reproducibility. This includes the post-hoc fine-tuning used for influence estimation and the sampling of coreset candidates. 

\section{Downstream Experiments Details}
\label{app: experiment details}

\subsection{RNA Downstream Tasks}

\subsubsection{RNA Type Classification (TypeCls)}

This task is to predict each RNA sequence into one of 10 functional RNA types (e.g., rRNA, tRNA, snoRNA)~\cite{amin2019evaluation}. We evaluate model performance using \textbf{accuracy} (overall correctness) and \textbf{F1 score} (robustness under class imbalance). A linear classification head is appended to the mean-pooled representation from the RNA-FM encoder, and all model parameters are fine-tuned using cross-entropy loss. The model is trained for 30 epochs with a batch size of 32 using the AdamW optimizer. We set the learning rate to $1 \times 10^{-4}$ and apply a weight decay of 0.01 throughout training.

\subsubsection{CRISPR On-Target Prediction (CRI-On)}

This task is to predict the editing efficiency of single-guide RNAs (sgRNAs) at target genomic loci~\cite{chuai2018deepcrispr}. Model performance is evaluated using the Mean Squared Error (MSE) and the Spearman correlation coefficient (SC). We append a linear regression head on top of the mean pooled representation from the RNA-FM encoder and fine-tune all model parameters using mean squared error (MSE) loss. The model is trained for 30 epochs with a batch size of 32 using the AdamW optimizer. We set the learning rate to $1 \times 10^{-4}$, and apply a weight decay of 0.01 throughout training.

\subsubsection{RNA Modification Prediction (Modif)}

This task is to predict the presence of 12 common RNA modifications for a given input sequence~\cite{duan2019dynamic}. Model performance is evaluated using the mean AUC (area under the ROC curve). We append a sigmoid-activated linear classification head on top of the mean pooled representation from the RNA-FM encoder and fine-tune all model parameters using binary cross-entropy loss. The model is trained for 30 epochs with a batch size of 32 using the AdamW optimizer. We set the learning rate to $5 \times 10^{-5}$, and apply a weight decay of 0.01 throughout training.

\subsubsection{Secondary Structure Prediction}

This task is to identify paired regions (stems) and unpaired regions (loops, bulges, junctions) within RNA molecules, using the bpRNA dataset~\cite{danaee2018bprna}. We evaluate model performance using the following metrics: 
Precision (positive predictive value), Recall (true positive rate), F1 score (harmonic mean of precision and recall), and Matthews Correlation Coefficient (MCC) (a balanced measure of binary classification quality even under class imbalance). 
We append a binary classification head to the mean-pooled representation from the RNA-FM encoder and fine-tune only the classification head using binary cross-entropy loss. The training procedure runs for 10 epochs with a batch size of 32, using the AdamW optimizer with a learning rate of $1 \times 10^{-5}$, and a weight decay of 0.01. All other configurations follow the RNA Type Classification setup.

\subsubsection{Contact Map Prediction}

This task evaluates the model’s ability to predict long-range structural contacts within RNA sequences, where a contact refers to a pair of nucleotides that are spatially proximal in 3D structure but distant in sequence. We use the dataset from \citet{chen2022rnafm} and report long-range top-$L$ and top-$0.5L$ precision metrics at multiple contact densities.
To perform this task, we append a binary classification head to the RNA-FM encoder and fine-tune only the task-specific head while keeping the encoder frozen, using binary cross-entropy loss. The model is trained for 10 epochs with a batch size of 32, using the AdamW optimizer (learning rate $1 \times 10^{-5}$, weight decay 0.01).

\subsubsection{Distance Map Prediction}

This task predicts pairwise Euclidean distances between nucleotides in the 3D structure of an RNA molecule. We use the dataset from \citet{chen2022rnafm} and evaluate performance with R$^2$, Spearman correlation (SC), Mean Absolute Error (MAE), and Mean Squared Error (MSE).
We append a regression head and fine-tune only the task head with the RNA-FM encoder frozen, using mean squared error loss. The training configuration matches the other tasks: 10 epochs, batch size 32, AdamW optimizer with a learning rate of $1 \times 10^{-5}$, and weight decay of 0.01.

\subsubsection{RBP-RNA Interaction Prediction}

This task aims to predict whether an RNA molecule interacts with a given RNA-binding protein (RBP), using the dataset from \citet{chen2022rnafm}.
We evaluate performance using accuracy, AUC, and AUPR. A binary classification head is added to the encoder output, and only this head is fine-tuned with the backbone frozen. Training is performed for 10 epochs, with a batch size of 32, using AdamW with a $1 \times 10^{-5}$ learning rate, and weight decay of 0.01.

\subsection{Protein Downstream Tasks}

\subsubsection{Binary Localization Prediction}

This task aims to predict whether a given protein sequence is localized to a specific cellular compartment, using the dataset from \citet{almagro2017deeploc}.  
We evaluate model performance using accuracy as the primary metric.  
To perform this task, we append a two-layer MLP classification head to the encoder output of ESM-C.  
During fine-tuning, we freeze the backbone and train only the classification head using cross-entropy loss.  
Training is conducted for 50 epochs with a batch size of 32, using the AdamW optimizer with a learning rate of $1 \times 10^{-5}$ and a weight decay of 0.01.

\subsubsection{Secondary Structure Prediction}

This task aims to predict the secondary structural class (e.g., alpha-helix, beta-strand, coil) of each residue in a protein sequence, based on the dataset provided by \citet{netsurfp}.  
We report accuracy as the primary evaluation metric, which reflects the per-residue classification correctness.
For this task, we append a two-layer token-level classification head to the ESM-C encoder.  
During fine-tuning, the backbone is frozen and only the classification head is trained using cross-entropy loss.  
Training is performed for 50 epochs with a batch size of 32, using the AdamW optimizer with a learning rate of $1 \times 10^{-5}$ and a weight decay of 0.01.

\subsubsection{PPI Affinity Prediction}

This task aims to predict the binding affinity between two interacting proteins, using the PDBbind dataset as processed by \citet{moal2012skempi}.  
We evaluate model performance using two standard regression metrics: Mean Absolute Error (MAE) and Root Mean Squared Error (RMSE), both of which quantify the deviation between predicted and ground-truth affinity values.
To perform this task, we append a two-layer regression head to the pooled representation of the protein pair, generated by the ESM-C encoder.  
The backbone is frozen during training, and only the regression head is optimized using mean squared error loss.  
Fine-tuning is conducted for 50 epochs with a batch size of 32, using the AdamW optimizer with a learning rate of $1 \times 10^{-5}$ and a weight decay of 0.01.

\subsection{Further Details}

For all downstream evaluation tasks, we report the average performance across three independent runs with random seeds set to 0, 1, and 2. 

\section{Derivations}
\label{appendix: derivation}

\subsection{The parameter change $\tilde{\theta}_{z_\text{tr}} - \tilde{\theta}$}
\label{appendix: sec: deriving the param change}

For completeness, we derive the parameter change $\tilde{\theta}_{z_\text{tr}} - \tilde{\theta}$ in the context of loss minimization.

According Assumption~\ref{assumption:subset-erm}, we first assume that \( \tilde{\theta} \) minimizes the empirical risk on $D_\text{sub}$:
\begin{equation}
    R_\text{sub} (\theta)
    := \frac{1}{M} \sum_{m=1}^M \ell(z_m, \theta),
\end{equation}
where $D_\text{sub} = \{ z_m \}_{m=1}^M \subset D_\text{tr}$.
We further assume that $R_\text{sub}$ is twice-differentiable and strongly convex in $\theta$, i.e.,
\begin{equation}
    \tilde{H}_\text{sub}
    := \nabla^2_{\theta} \hat{R}_\text{sub} (\tilde{\theta})
    = \frac{1}{M} \sum_{m=1}^M \nabla^2_{\tilde{\theta}} \ell(z_m, \tilde{\theta}),
\end{equation}
exists and is positive definite. This guarantees the existence of $\tilde{H}_\text{sub}^{-1}$, which we will use in the subsequent derivation.

The perturbed parameters $\tilde{\theta}_{z_\text{tr}}$ can be written as
\begin{equation}
    \label{appendix: deriving: parameter change}
    \tilde{\theta}_{z_\text{tr}}
    := \arg \min_\theta R_\text{sub}(\theta) + \epsilon \ell(z_\text{tr}, \theta).
\end{equation}

Since $\tilde{\theta}_{z_\text{tr}}$ is a minimizer of \Cref{appendix: deriving: parameter change}, let us examine its first-order optimality conditions:
\begin{equation}
    0 = \nabla_{\tilde{\theta}} R_\text{sub}(\tilde{\theta}_{z_\text{tr}}) + \epsilon \nabla_{\tilde{\theta}} \ell(z, \tilde{\theta}_{z_\text{tr}})
\end{equation}

Next, since $\tilde{\theta}_{z_\text{tr}} \to \tilde{\theta}$ as $\epsilon \to 0$, we perform a Taylor expansion of the right-hand side:
\begin{equation}
    0
    \approx \left[ \nabla_{\tilde{\theta}} R_\text{sub}(\tilde{\theta}) + \epsilon \nabla_{\tilde{\theta}} \ell (z_\text{tr}, \tilde{\theta}) \right]
    + \left[ \nabla^2_{\tilde{\theta}} R_\text{sub}(\tilde{\theta}) + \epsilon \nabla^2_{\tilde{\theta}} \ell(z_\text{tr}, \tilde{\theta}) \right] \left( \tilde{\theta}_{z_\text{tr}} - \tilde{\theta} \right),
\end{equation}
where we have dropped $o(\| \tilde{\theta}_{z_\text{tr}} - \tilde{\theta} \|)$ terms.

Sovling for $\tilde{\theta}_{z_\text{tr}} - \tilde{\theta}$, we get:
\begin{equation}
    \tilde{\theta}_{z_\text{tr}} - \tilde{\theta}
    \approx - \left[ \nabla^2_{\tilde{\theta}} R_\text{sub}(\tilde{\theta}) + \epsilon \nabla^2_{\tilde{\theta}} \ell(z_\text{tr}, \tilde{\theta}) \right]^{-1}
    \left[ \nabla_{\tilde{\theta}} R_\text{sub}(\tilde{\theta}) + \epsilon \nabla_{\tilde{\theta}} \ell (z_\text{tr}, \tilde{\theta}) \right].
\end{equation}
Since $\tilde{\theta}$ is the minimizer of $\hat{R}^\gamma_\text{sub}(\theta)$, we have $\nabla_{\tilde{\theta}} R_\text{sub}(\tilde{\theta}) = 0$.
Dropping $o(\epsilon)$ terms, then
\begin{equation}
    \tilde{\theta}_z - \tilde{\theta}
    \approx - \epsilon \tilde{H}^{-1}_\text{sub} \tilde{g}_{z_\text{tr}},
\end{equation}
where $\tilde{g}_{z_\text{tr}} = \nabla_{\tilde{\theta}} \ell (z_\text{tr}, \tilde{\theta})$.

\subsection{The Influence Function}
\label{appendix: sec: deriving the influence function}

Recall \Cref{eq: risk change on tilde theta}, we have
\begin{equation}
    \ell(z_\text{sub}, \tilde{\theta}_{z_\text{tr}}) - \ell(z_\text{sub}, \tilde{\theta})
    \approx
    \tilde{g}_{z_\text{sub}}^\top \Delta \theta +
    \frac{1}{2} \Delta \theta^\top \nabla^2_{\tilde{\theta}} \ell (z_{\text{sub}}, \tilde{\theta}) \Delta \theta,
\end{equation}
Combining with \Cref{eq: parameter change on tilde theta}, we have
\begin{align}
    \ell(z_\text{sub}, \tilde{\theta}_{z_\text{tr}}) - \ell(z_\text{sub}, \tilde{\theta})
    & \approx
    \tilde{g}_{z_\text{sub}}^\top \left( - \epsilon \tilde{H}^{-1}_\text{sub} \tilde{g}_{z_\text{tr}} \right) +
    \frac{1}{2} \left( - \epsilon \tilde{H}^{-1}_\text{sub} \tilde{g}_{z_\text{tr}} \right)^\top \nabla^2_{\tilde{\theta}} \ell (z_{\text{sub}}, \tilde{\theta}) \left( - \epsilon \tilde{H}^{-1}_\text{sub} \tilde{g}_{z_\text{tr}} \right) \\
    & =
    - \epsilon \tilde{g}_{z_\text{sub}}^\top \tilde{H}^{-1}_\text{sub} \tilde{g}_{z_\text{tr}}+
    \frac{1}{2} \epsilon^2 \tilde{g}_{z_\text{tr}}^\top \tilde{H}^{-1}_\text{sub} \nabla^2_{\tilde{\theta}} \ell (z_{\text{sub}}, \tilde{\theta}) \tilde{H}^{-1}_\text{sub} \tilde{g}_{z_\text{tr}}.
\end{align}
Since we want to measure the loss change with respect to removing the training sample $z_\text{tr}$, we define the Influence Function as follows:
\begin{equation}
    \label{eq: influence on single test sample}
    \mathcal{I}(z_\text{tr}, z_\text{sub})
    :=
    \tilde{g}_{z_\text{sub}} \tilde{H}^{-1}_\text{sub} \tilde{g}_{z_\text{tr}}
    + \frac{1}{2} \epsilon \tilde{g}_{z_\text{tr}}^\top \tilde{H}^{-1}_\text{sub} \nabla^2_{\tilde{\theta}} \ell (z_{\text{sub}}, \tilde{\theta}) \tilde{H}^{-1}_\text{sub} \tilde{g}_{z_\text{tr}},
\end{equation}
with $\epsilon > 0$.
Next, we consider the influence on the subset $D_\text{sub}$, which can be defined as the sum of the influence on each subset sample:
\begin{equation}
    \mathcal{I}(z_\text{tr}, D_\text{sub})
    := \sum_{m=1}^{M} \mathcal{I}(z_\text{tr}, z_m).
\end{equation}
Incorporating with \Cref{eq: influence on single test sample}, we get
\begin{equation}
    \begin{aligned}
        \mathcal{I}(z_\text{tr}, D_\text{sub})
        & := \sum_{m=1}^M
        \tilde{g}_{z_m} \tilde{H}^{-1}_\text{sub} \tilde{g}_{z_\text{tr}}
        + \frac{1}{2} \epsilon \tilde{g}_{z_\text{tr}}^\top \tilde{H}^{-1}_\text{sub} \nabla^2_{\tilde{\theta}} \ell (z_{\text{sub}}, \tilde{\theta}) \tilde{H}^{-1}_\text{sub} \tilde{g}_{z_\text{tr}} \\
        & = \tilde{g}_\text{sub} \tilde{H}^{-1}_\text{sub} \tilde{g}_{z_\text{tr}}
        + \frac{1}{2} \epsilon \tilde{g}_{z_\text{tr}}^\top \tilde{H}^{-1}_\text{sub} \tilde{g}_{z_\text{tr}}.
    \end{aligned}
\end{equation}
Since $\tilde{g}_\text{sub} \to 0$, we define the Influence Function on the subidation set $D_\text{sub}$ as follows:
\begin{equation}
    \mathcal{I}(z_\text{tr}, D_\text{sub})
    := \tilde{g}_{z_\text{tr}}^\top \tilde{H}^{-1}_\text{sub} \tilde{g}_{z_\text{tr}},
\end{equation}
where $\frac{1}{2} \epsilon > 0$ is dropped.

\end{document}